\documentclass[10pt,twocolumn,letterpaper]{article}

\usepackage[pagenumbers]{cvpr} 

\definecolor{cvprblue}{rgb}{0.21,0.49,0.74}
\usepackage[pagebackref,breaklinks,colorlinks,allcolors=cvprblue]{hyperref}
\usepackage{gensymb}
\def\paperID{16931} 
\def\confName{CVPR}
\def\confYear{2026}

\title{Volumetrically Consistent Implicit Atlas Learning via Neural Diffeomorphic Flow for Placenta MRI}

\author{Athena Taymourtash$^1$\and S. Mazdak Abulnaga$^{1,2}$\and Esra Abaci Turk$^3$\and P. Ellen Grant$^3$\and Polina Golland$^1$ \\
$^1$\normalsize{MIT Computer Science and Artificial Intelligence Laboratory} \\ 
$^2$\normalsize{Massachusetts General Hospital, Harvard Medical School} \\ $^3$\normalsize{Boston Children’s Hospital, Harvard Medical School} \\
{\tt\small taymourtash@csail.mit.edu}
}

\begin{document}
\maketitle
\begin{abstract}
Establishing dense volumetric correspondences across anatomical shapes is essential for group-level analysis but remains challenging for implicit neural representations. Most existing implicit registration methods rely on supervision near the zero-level set and thus capture only surface correspondences, leaving interior deformations under-constrained.
We introduce a volumetrically consistent implicit model that couples reconstruction of signed distance functions (SDFs) with neural diffeomorphic flow to learn a shared canonical template of the placenta.
Volumetric regularization, including Jacobian-determinant and biharmonic penalties, suppresses local folding and promotes globally coherent deformations. In the motivating application to placenta MRI, our formulation jointly reconstructs individual placentas, aligns them to a population-derived implicit template, and enables voxel-wise intensity mapping in a unified canonical space.
Experiments on in-vivo placenta MRI scans demonstrate improved geometric fidelity and volumetric alignment over surface-based implicit baseline methods, yielding anatomically interpretable and topologically consistent flattening suitable for group analysis. 
\end{abstract}    
\section{Introduction}
\label{sec:intro}
Representing and comparing 3D anatomical shapes across individuals is fundamental in many medical imaging tasks, including segmentation~\cite{you2020unsupervised, tang2021recurrent, yan2022after, you2022simcvd, khan2022implicit}, reconstruction~\cite{you2020unsupervised, lebrat2021corticalflow}, surgical navigation~\cite{lehmann2009integrating, suinesiaputra2017statistical}, and diagnostic pathology~\cite{you2020unsupervised, elkassem2022multiinstitutional}. The placenta, a vital lifeline organ whose normal functioning is essential for fetal growth~\cite{abulnaga2021volumetric}, presents particular challenges: its in-vivo shape conforms to the curved uterine wall at arbitrary locations and orientation, varies widely across subjects, and changes dynamically due to fetal motion, maternal breathing, and contractions during imaging~\cite{abulnaga2021volumetric}. Unlike other organs with established atlases, the placenta lacks a standardized geometric representation. To enable population-level comparisons, one must model placental shapes with anatomical fidelity while also learning transformations that map one highly deformed placenta to another. 

Existing methods for anatomical normalization of the placenta have focused on explicit geometric flattening into a canonical plane. Early work flattened the 2D layers independently, leading to distortion and misalignment across depth~\cite{miao2017placenta}. 
A recent solution employed a volumetric parameterization based on tetrahedral meshes to achieve locally injective flattening~\cite{abulnaga2021volumetric,abulnaga2019placental}. While this formulation improves geometric fidelity, it must be solved separately for each subject, depends on careful initialization, and does not provide dense correspondences or a common template across a population. Moreover, its reliance on explicit meshes and iterative optimization limits scalability and generalization.

Recent advances in deep implicit functions (DIF) have demonstrated remarkable capability in representing complex shapes and fine geometric details~\cite{park2019deepsdf,mescheder2019occupancy,chen2019learning,saito2019pifu,xu2019disn,gropp2020implicit,deng2021deformed,Sun_2022_CVPR}. Compared to explicit representations based on voxel grids, point clouds, and meshes, DIFs are compact, resolution-agnostic, and differentiable, which makes them suitable for end-to-end training. Despite these advantages, conventional DIFs do not establish correspondences across instances; each shape is represented independently by its own scalar field. This limitation precludes population-level shape analysis and downstream applications such as label transfer, which depend on consistent coordinate mappings between instances.
To address this, subsequent extensions have introduced template-based implicit representations that embed all shapes into a shared latent or canonical space through learned deformation fields~\cite{zheng2021deep,deng2021deformed,yang2022implicitatlas, Sun_2022_CVPR}.
These models jointly learn to reconstruct each shape’s implicit field and align it to a shared canonical template that emerges during training in a single end-to-end process.

While these approaches enable dense surface correspondences, their supervision remains effectively surface-biased. Although losses are computed across the domain, the SDF loss gradient magnitude decays rapidly with distance from the zero-level set, leaving the deformation field under-constrained inside the volume. As a result, interior trajectories may fold or drift, compromising diffeomorphic continuity and volumetric coherence.
Such behavior is particularly problematic in medical applications like placenta analysis, where analysis relies on consistent mapping of volumetric MRI intensities rather than solely surface geometry.

In this work, we propose a volumetrically consistent implicit model that extends surface-based implicit registration to the full organ volume. Our method learns diffeomorphic mappings that jointly reconstruct individual placentas, align them to a shared implicit template, and flatten them into a canonical configuration within a single end-to-end process. 
To achieve spatially consistent volumetric warps, we enforce regularization that discourages local folding and promotes globally smooth, low-distortion deformations throughout the tissue volume.
The resulting model produces high-fidelity reconstructions and a biologically interpretable flattened template that emerges naturally during training. This volumetric formulation bridges implicit shape modeling and diffeomorphic registration, providing a unified framework for anatomically faithful atlas construction and population-level analysis.

\section{Related Work}
\label{sec:rw}
\noindent\textbf{Deep Implicit Functions.} Implicit functions have become an efficient tool for continuous and differentiable shape modeling. Instead of discretizing 3D geometries as meshes or voxels, these methods learn functions $f_\theta(\mathbf{x})$ to map spatial coordinates to a continuous scalar field such as signed distance~\cite{park2019deepsdf} or occupancy probability~\cite{mescheder2019occupancy, chen2019learning}. The shape surface is then implicitly expressed as the isocontour of $f_\theta$, and reconstructed with, for example, Marching Cubes~\cite{lorensen1998marching} at arbitrary resolution. Regularizers on Eikonal or surface normals improve geometric fidelity near the zero-level set~\cite{gropp2020implicit}. However, these models are \emph{instance-wise}: each object’s field is learned independently, yielding no cross-instance correspondence. To address this, template-based methods decouple geometry and deformation, learning a canonical field and a conditional warp that maps subject coordinates into the template space~\cite{zheng2021deep,deng2021deformed,Sun_2022_CVPR,yang2022implicitatlas}. Since the training signals remain concentrated near the zero set due to SDF/occupancy losses, the volumetric behavior of the resulting reconstruction is largely uncontrolled. In practice this produces surface-accurate but interior-ambiguous warps (e.g., drifting trajectories, local folding), limiting intensity transfer and population analyses that require coherent mappings throughout the organ volume.

\noindent\textbf{Diffeomorphic Transformation.}
Diffeomorphic mapping provides a mathematically grounded way to model smooth, invertibe, and topology-preserving deformations between shapes~\cite{beg2005computing,vercauteren2009diffeomorphic,avants2008symmetric,ashburner2007fast}. 
Integrating stationary or time-varying velocity fields (SVF/TVF) with geodesic constraints and log-domain parameterizations provides numerically stable and well-behaved flows. Recent learning-based variants, including VoxelMorph~\cite{balakrishnan2019voxelmorph}, DDF-Net~\cite{dalca2019unsupervised}, and neural velocity field based methods~\cite{han2023diffeomorphic, mok2020fast}, have adapted these principles to deep architectures for dense medical registration. 

More recently, Neural ODE~\cite{chen2018neural}  has enabled continuous and invertible warps parameterized by neural velocity fields, offering memory efficient integration and analytical control of the deformation Jacobian. Such neural diffeomorphic flows have been successfully applied to shape correspondence~\cite{gupta2020neural,Sun_2022_CVPR,han2024hybrid}, 3D reconstruction, articulated body pose estimation, dynamic scene modeling, and yet they are typically defined over voxel grids or surface point sets. 
Our approach adopts this continuous-time formalism but operates directly on the implicit function space, producing volumetric, topology-preserving transformations that guarantee smoothness and invertibility within the organ interior.

\noindent\textbf{Atlases and Dense Correspondences.}
Anatomical atlases aim to establish a common coordinate system that represents population variability while preserving subject-specific anatomy. Classical methods such as ANTs~\cite{avants2009advanced} and SPM~\cite{penny2011statistical} build brain atlases through iterative registration and intensity averaging, often relying on diffeomorphic velocity fields~\cite{beg2005computing}.
Recent deep learning approaches have accelerated this process by learning parametric mappings to a shared template space~\cite{dalca2019unsupervised,balakrishnan2019voxelmorph}, or by framing atlas construction as a test-time adaptation or groupwise registration task~\cite{abulnaga2025multimorph,he2023groupwise}.
These methods enable efficient atlas estimation without repeated optimization, but typically operate on voxel intensities and therefore implicitly assume limited shape variability across subjects. For anatomies with substantial geometric variability, such as the placenta, intensity-only registration often fails to establish stable correspondences~\cite{abulnaga2021volumetric, turk2020placental}, motivating approaches that model shape geometry. 

In shape analysis, implicit template models such as DIT~\cite{zheng2021deep} and deformation-aware autoencoders~\cite{Sun_2022_CVPR,le2023second} have begun to unify correspondence learning and template discovery in a single differentiable framework, aiming to establish continuous, invertible population representations beyond voxel-based formulations.
Nevertheless, scalability and geometric fidelity remain open challenges.



\section{Methods}
\label{sec:methods}
\begin{figure*}[t]
  \centering
  \includegraphics[width=1\linewidth]{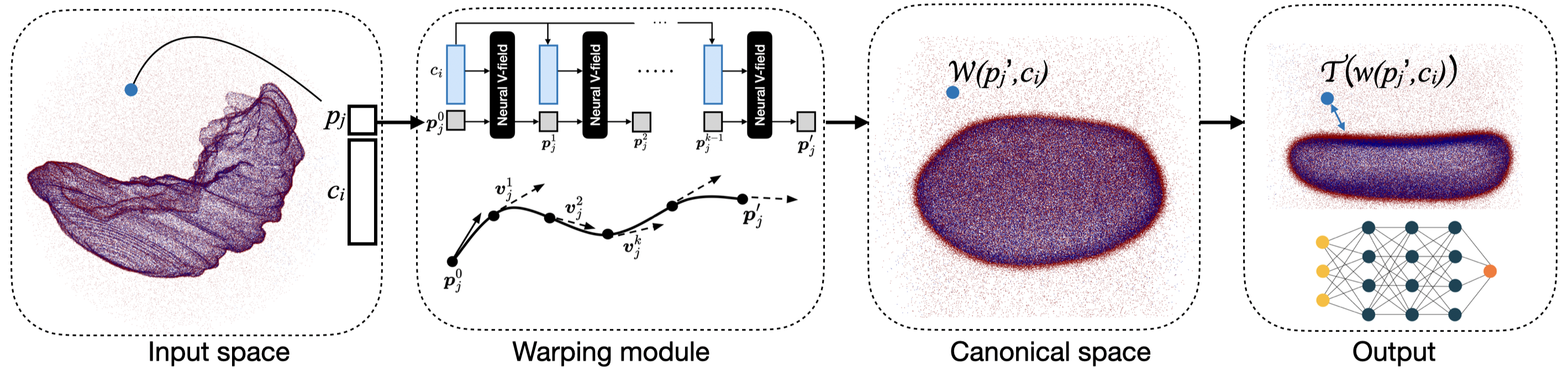}

  \caption{Method overview. The warping function transforms point samples of shape $i$ to their canonical positions, which are then mapped to SDF values by the implicit template.}
  \label{fig:twocol}
\end{figure*}
Our approach follows the formulation of Deep Implicit Templates (DIT)~\cite{zheng2021deep}, which decomposes each encoded shape into a shared template SDF and a conditional spatial deformation. Building on this idea, we model the deformation as a conditional diffeomorphic flow~\cite{gupta2020neural} defined over volumetric tetrahedral embeddings, ensuring smooth, topology-preserving mappings that remain consistent throughout the interior of the placenta.


\subsection{Review of Deep Implicit Template}
Deep Implicit Templates (DIT)~\cite{zheng2021deep} represent a collection of shapes $\left\{X_i\right\}_{i=1}^N$ through a conditional signed distance function:
\begin{equation}
    \mathcal{F}:\mathbb{R}^3 \times \mathcal{G} \to \mathbb{R}, \quad \mathcal{F}(\boldsymbol{x}; \boldsymbol{c_i})=s,
\end{equation}
where $x$ is a spatial coordinate, $c_i \in \mathcal{G}$ is the latent shape code, and $s$ is the predicted signed distance to the closest surface point. Unlike DeepSDF~\cite{park2019deepsdf}, where the latent code jointly encodes geometry and pose, DIT explicitly separates these factors by introducing a template function $\mathcal{T}_\theta: \mathbb{R}^3\mapsto \mathbb{R}$ that defines a single shared implicit surface across the population, and a conditional warp $\mathcal{W}_\psi(.;c_i): \mathbb{R}^3 \times \mathcal{G} \to \mathbb{R}^3$ that maps coordinates from an instance’s space to a canonical template space:
\begin{equation}
     \mathcal{F}(\boldsymbol{x}; \boldsymbol{c_i})=\mathcal{T}_\theta(\mathcal{W}_\psi(\boldsymbol{x}; \boldsymbol{c_i})) 
\end{equation}
This decomposition enables implicit correspondence between shapes via the shared template but does not enforce $\mathcal{W}_\psi$ to be smooth or invertible, often resulting in local distortions or topological inconsistencies. 

During inference, a new shape is represented by optimizing its latent code:
\begin{equation}
    \hat{\boldsymbol{c}}_i=\underset{\boldsymbol{c}_i}{\arg \min } \sum_{(\boldsymbol{x_j}, s_j) \in X_i} \ell\left(\mathcal{T}_\theta\left(\mathcal{W}_\psi\left(\boldsymbol{x}; \boldsymbol{c}_i\right), s_j\right)\right)+\frac{1}{\sigma^2}\left\|\boldsymbol{c}_i\right\|_2^2
\end{equation}
where $\ell(.,.)$ denotes an $L_1$ reconstruction loss between predicted and ground-truth SDF values, while the template and deformation networks $\left(\mathcal{T}_\theta, \mathcal{W}_\psi\right)$ remain fixed.

\subsection{Neural Diffeomorphic Flow}
Let $\Phi(\boldsymbol{x},t;c_i)$ denote the continuous flow trajectory of a 3D coordinate $\boldsymbol{x}$ for instance $i$ governed by a latent-conditioned velocity field $v(.,t; c_i)$. For clarity of notation, we write $\Phi_i(\boldsymbol{x},t)$ as shorthand for 
$\Phi(\boldsymbol{x},t;c_i)$: 
\begin{equation}\label{eq:4}
    \frac{\partial \Phi_i(\boldsymbol{x}, t)}{\partial t}=v\left(\Phi_i(\boldsymbol{x}, t), t;c_i\right), \quad \Phi_i(\boldsymbol{x}, 0)=\boldsymbol{x}
\end{equation}
where $v\left(\cdot, t; c_i\right):\mathbb{R}^3 \times [0,1] \to \mathbb{R}^3$ denotes a neural velocity field conditioned on the latent code $c_i$ and time $t$. Integrating Eq.(\ref{eq:4}) from $t=0$ to $t=1$ yields the forward deformation:
\begin{equation}
    \Phi_i\left(\boldsymbol{x},1\right)=x +\int_0^1{v\left(\Phi_i\left(\boldsymbol{x},t\right),t;c_i\right)dt,}
\end{equation}
which defines the mapping from the instance’s coordinate space to the canonical template domain. The predicted signed distance for coordinate $\boldsymbol{x}$ of instance $i$ is then obtained by evaluating the canonical template field at the mapped coordinate as $s_i(\boldsymbol{x})=\mathcal{T}_\theta\left(\Phi_i\left(\boldsymbol{x},1\right)\right)$. If $v$ is Lipschitz-continuous, the solution of Eq.(\ref{eq:4}) is unique and defines a diffeomorphic mapping~\cite{gupta2020neural}.

The inverse deformation $\Psi(\boldsymbol{x},t,c_i)$ is defined by integrating the backward flow:
\begin{equation}
    \frac{\partial \Psi(\boldsymbol{x}, t;c_i)}{\partial t}=-v\left(\Psi(\boldsymbol{x}, t;c_i), t; c_i\right), \Psi(\boldsymbol{x}, 0;c_i)=\boldsymbol{x}, \quad   
\end{equation}
\begin{equation}
    \Phi_i^{-1}(\boldsymbol{x},0)=\Psi(\boldsymbol{x}, 1, c_i),
\end{equation}
which guarantees invertibility and allows consistent correspondences in both surface and volumetric domains.

Following~\cite{Sun_2022_CVPR}, we approximate the time-dependent velocity field as a composition of $K$ stationary sub-fields: 
\begin{equation}
    v\left(\Phi_i(x, t), t; c_i\right)=\sum_{k=1}^K \chi_{\left[\frac{k-1}{K}, \frac{k}{K}\right)}(t) v^{(k)}\left(\Phi_i(x, t);c_i\right),
\end{equation}
where $\chi_A$ is the indicator function. This discretized representation corresponds to a sequence of piecewise-stationary velocity updates, each modeled by a neural-ODE (NODE) block~\cite{gupta2020neural} conditioned on the latent code $c_i$.

Although $\Phi_i(.,t)$ is defined over the full $\mathbb{R}^3$ domain, the supervision signal from the SDF reconstruction loss is predominantly concentrated near the zero-level set. To impose volumetric regularity, we introduce a tetrahedral embedding $\mathcal{M}_i=(V_i, K_i)$ for each instance, where $V_i$ are tetrahedral vertices and $K_i$ is the connectivity set, and explicitly evaluate the flow on $V_i$ during training. These tetrahedra are not used as a geometric representation; they merely provide interior sample points at which the deformation field $\Phi_i$ is queried.  
These volumetric evaluations serve as geometric constraints, enforcing smooth, topology-preserving trajectories inside the placenta volume.

\subsection{Network Training}
All parameters of the deformation function $\mathcal{W}_\psi$, the template SDF $\mathcal{T}_\theta$, and latent codes $\left\{c_i\right\}$ are optimized jointly over all subjects using the total loss:
\begin{equation}\label{eq:9}
    \mathcal{L} = \mathcal{L}_{\mathrm{rec}}+\lambda_{\mathrm{d}}\mathcal{L}_{\mathrm{d}}+\lambda_{\mathrm{Jac}}\mathcal{L}_{\mathrm{Jac}}+ \lambda_{\mathrm{biH}}\mathcal{L}_{\mathrm{biH}}+\sum_{i=1}^N\left\|\boldsymbol{c}_i\right\|_2^2
\end{equation}
\noindent\textbf{Reconstruction loss.} At each iteration, the signed-distance reconstruction loss is evaluated at multiple time steps $t \in \left\{0.25, 0.5, 0.75, 1\right\}$ along the deformation trajectory: 
\begin{equation}
    \mathcal{L}_{r e c}=\sum_{t \in T}\sum_{i=1}^N \sum_{j=1}^M \mathcal{L}_{\epsilon_t, \lambda_t}\left(\mathcal{T}_\theta\left(\Phi_i\left({x_j,t}\right)\right), s_{i,j}\right),
\end{equation}
where $x_j$ is a spatial sample, $\Phi_i({x_j,t})$ its deformation at time $t$, and $s_{i,j}$ the ground-truth SDF value. $\mathcal{L}_{\epsilon_t, \lambda_t}$ denotes a soft $L_1$ loss with tolerance $\epsilon_t$ and weighting factor $\lambda_t$ controlling the emphasis on hard examples~\cite{zheng2021deep, duan2020curriculum}. A temporal curriculum is realized by varing these parameters across time: early deformations use larger $\epsilon_t$ and smaller $\lambda_t$ for coarse alignment, while later steps adopt smaller $\epsilon_t$ and larger $\lambda_t$ for finer geometric accuracy.

\noindent\textbf{Deformation Regularization.} As all shapes are normalized to the unit cube and approximately aligned, we regularize the displacement magnitude to prevent excessive warps:
\begin{equation}
    \mathcal{L}_{\mathrm{d}}=\sum_{t \in T}\sum_{i=1}^N \sum_{j=1}^M h\left(\left\|\Phi\left(\boldsymbol{x}_j,t; \boldsymbol{c}_i\right)-\boldsymbol{x}_j\right\|_2\right), 
\end{equation}
where $h(\cdot)$ is the Huber loss with $\delta = 0.25$.

\noindent\textbf{Volumetric Regularization.}
While the SDF loss provides strong supervision near the surface, gradients decay rapidly away from the zero-level set, leaving the interior deformation under-constrained.
To promote volumetric regularity, we evaluate $\Phi_i(\boldsymbol{x})$ on the interior tetrahedral samples $x \in \Omega_i$ drawn from the embedding $\mathcal{M}_i=\left(V_i,K_i\right)$. Two penalties are applied: 
\begin{equation}
    \mathcal{L}_{\mathrm{Jac}}=\sum_{t \in T}\mathbb{E}_{x \in \Omega_i}\left[\max \left(0,-\operatorname{det} \nabla \Phi_i(\boldsymbol{x}, t)\right)\right]^2,
\end{equation}
\begin{equation}
    \mathcal{L}_{\mathrm{biH}}=\sum_{t \in T}\mathbb{E}_{x \in \Omega_i}\left[\left\|\Delta^2 \Phi_i(\boldsymbol{x}, t)\right\|_2^2\right],
\end{equation}
where $\mathcal{L}_{\mathrm{Jac}}$ penalizes local folding $\left(\operatorname{det} \nabla \Phi_i<0\right)$, and $\mathcal{L}_{\mathrm{biH}}$ applies the biharmonic operator $\Delta^2$ to the deformation field to enforce globally smooth, low-distortion volumetric trajectories.

\subsection{Intensity Mapping}
Prior to training, each placenta surface is rigidly normalized to the unit cube and reoriented such that its first principal component aligns with the canonical z-axis, defining a composed transform $A_i:\mathbb{R}^3 \to \mathbb{R}^3$ from scnner space to normalized coordinate.

Given the learned diffeomorphic flow $\Phi_i$, voxel-wise MRI intensities are transferred from each instance image $I_i(x)$ to the canonical template space by pullback through the inverse mapping:
\begin{equation}
I_T(x_T) = I_i(A_i^{-1}\Phi_i^{-1}(x_T)),
\label{eq:intensity_pullback}
\end{equation}
where $x_T$ denotes a voxel center in the template domain. The inverse flow $\Phi_i^{-1}$ is obtained by integrating the velocity field backward in time, and intensities are sampled from $I_i$ using trilinear interpolation at non-grid coordinates.

For voxels inside the placenta, the deformation is evaluated on the tetrahedral embedding $\mathcal{M}_i = (V_i, K_i)$. 
Each template-space point $x_T$ lies in a tetrahedron $k \in K_i$ with vertex coordinates $X_k=[X_k^{(1)},...,X_k^{(4)}] \in \mathbb{R}^{3\times4}$ and corresponding instance-space vertices $Z_k = \Phi_i^{-1}(X_k)$. Let $B=[X_k^{(2)}-X_k^{(1)},X_k^{(3)}-X_k^{(2)},X_k^{(4)}-X_k^{(1)}]$ denote the edge matrix. The barycentric coefficients $\alpha$ of $x_T$ satisfy:

\begin{equation}
[\alpha_2,\alpha_3,\alpha_4]^{\top}
= (X_k B)^{-\top} (x_T - X_k^{(1)}),
\label{eq:barycentric}
\end{equation}
\begin{equation}
    \alpha_1=1-\sum_{j=2}^4\alpha_j
\end{equation}
and the mapped coordinate in the subject space is obtained as:
\begin{equation}
z = Z_k \alpha, \quad I_T\left(x_T\right)=I_i\left(z\right),
\label{eq:mapped_coord}
\end{equation}
This piecewise-affine barycentric formulation ensures an injective, locally continuous mapping of voxel intensities across the placenta volume.

\section{Experiments}
\label{sec:expr}

\begin{table*}[!t]
\centering
\small 
\setlength{\tabcolsep}{3.8pt}
\renewcommand{\arraystretch}{1.1}
\caption{
Surface accuracy metrics on the placenta dataset. CD: Chamfer Distance (multiplied by $10^3$), EMD: Earth Mover’s Distance, and NC: Normal Consistency.
Precision, Recall, and $\mathrm{F}_1$ are evaluated at distance thresholds of 0.01, 0.03, and 0.05. 
}
\begin{tabular}{lcccccccccccc}
\toprule
& Chamfer-L2 $(\downarrow)$ & EMD $(\downarrow)$ & NC $(\uparrow)$ 
& \multicolumn{3}{c}{Precision $(\uparrow)$} 
& \multicolumn{3}{c}{Recall $(\uparrow)$} 
& \multicolumn{3}{c}{$\mathrm{F}_1$ $(\uparrow)$} \\ 
\cmidrule(lr){5-7} \cmidrule(lr){8-10} \cmidrule(lr){11-13}
Model & & & & @0.01 & @0.03 & @0.05 & @0.01 & @0.03 & @0.05 & @0.01 & @0.03 & @0.05 \\ 
\midrule
DeepSDF~\cite{park2019deepsdf} & 0.27 & 0.05 & 0.97 & 0.24 & 0.98 & 0.99 & 0.24 & 0.99 & \textbf{1.00} & 0.24 & 0.99 & 0.99 \\
DIT~\cite{zheng2021deep}       & 0.17 & 0.05 & 0.97 & 0.37 & 0.98 & 0.99 & 0.37 & 0.99 & \textbf{1.00} & 0.38 & 0.99 & 0.99 \\
DIF-Net~\cite{deng2021deformed}       & 0.12 & \textbf{0.04} & \textbf{0.98}  & 0.43 & 0.99 & \textbf{1.00} & 0.43 & 0.99 &  \textbf{1.00}& 0.43 & 0.99 & \textbf{1.00} \\
NDF~\cite{Sun_2022_CVPR}       & 0.22 & 0.05 & 0.97 & 0.28 & \textbf{0.99} & \textbf{1.00} & 0.28 & \textbf{0.99} & \textbf{1.00} & 0.28 & \textbf{0.99} & \textbf{1.00} \\
\midrule
\textbf{Ours}                  & \textbf{0.12} & \textbf{0.04} & \textbf{0.98} & \textbf{0.43} & 0.99 & \textbf{1.00} & \textbf{0.43} & \textbf{0.99} & \textbf{1.00} & \textbf{0.43} & 0.99 & \textbf{1.00} \\
\bottomrule
\end{tabular}
\label{tab:surf_accuracy}
\end{table*}
\paragraph{Dataset.}
We trained and evaluated our method, as well as all baseline approaches, on 111 MRI scans acquired from 78 pregnancies (26 \text{--} 38 weeks of gestation), including 60 singleton and 18 twin cases.
For 33 singleton subjects, scans were acquired in both supine and left lateral maternal positions, yielding 66 distinct placenta shapes that were treated as independent training instances.
Each manually segmented placenta volume was converted to a surface mesh using the Marching Cubes algorithm, normalized to the unit cube, and reoriented such that the first principal axis aligned with the \textit{z}-axis, defining the normalization transform $A_i$. Following DeepSDF~\cite{park2019deepsdf}, we sampled normalized signed distance function (SDF) points to supervise implicit reconstruction.

\noindent\textbf{Experimental Setup.}
We compare our method against representative implicit deformation and template-learning approaches, including DeepSDF~\cite{park2019deepsdf}, DIF-Net~\cite{deng2021deformed}, DIT~\cite{zheng2021deep}, and NDF~\cite{Sun_2022_CVPR}.
All baseline methods were trained using identical SDF samples and normalization transforms $A_i$ to ensure fair comparisons. For evaluation, we reconstruct each instances’s surface and volumetric mesh from the predicted SDF field and compute geometric, diffeomorphic, and distortion metrics to assess both surface and internal regularity during deformation, and detect folding or distortion within the organ.

Our network follows the DIT~\cite{zheng2021deep} architecture with an unconditional decoder and a four-step Neural ODE warper conditioned on 256-D latent codes. We use the Adam optimizer with an initial learning rate of $10^{-3}$, decayed by a factor of 0.5 every 500 epochs, and a batch size of 8. Training was performed on NVIDIA RTX~A6000 GPUs for 2000 iterations. Additional details on the architecture and training are provided in the supplementary material.

\subsection{Surface Accuracy}
Table~\ref{tab:surf_accuracy} reports surface reconstruction accuracy for all methods.
All methods achieve sub-voxel Chamfer and EMD errors with high normal consistency, indicating that the overall placental geometry is well constrained by the SDF supervision. Precision measures the fraction of predicted surface points that are correct (low false positives), recall measures the fraction of the ground-truth surface that is recovered (low false negatives), and the 
$\mathrm{F}_1$ score is their harmonic mean, providing a balanced measure of surface 
localization accuracy.

\begin{table}[t]
\centering
\footnotesize 
\setlength{\tabcolsep}{1.2pt}
\renewcommand{\arraystretch}{1.1}
\caption{Diffeomorphism metrics computed on tetrahedral embeddings. 
FlipRate measures the percentage of tetrahedra with negative signed volume, logDet–$\mathrm{L}_1$ captures the average deviation of local volume change from unity, and CycleError quantifies the inverse-consistency of the forward and backward flows. All values are averaged across subjects; lower is better.}
\begin{tabular}{lccc}
\toprule
 & FlipRate$(\downarrow)$ & logDet-$\mathrm{L}_1$$(\downarrow)$& CycleError$(\downarrow)$ \\
\midrule
DIF-Net~\cite{deng2021deformed}             & 15.34 & 0.81 & 0.24 \\
DIT~\cite{zheng2021deep}                   & 11.88 & \textbf{0.67} &  0.29\\
NDF~\cite{Sun_2022_CVPR}                   & 8.78 & 0.90 & 0.26 \\
\midrule
Ours                                       & \textbf{4.04} & 0.69 & \textbf{0.20} \\
\bottomrule
\end{tabular}
\label{tab:diff}
\end{table}

\begin{figure}[b]
  \centering
  \vspace{-4pt}
  \includegraphics[width=0.8\linewidth]{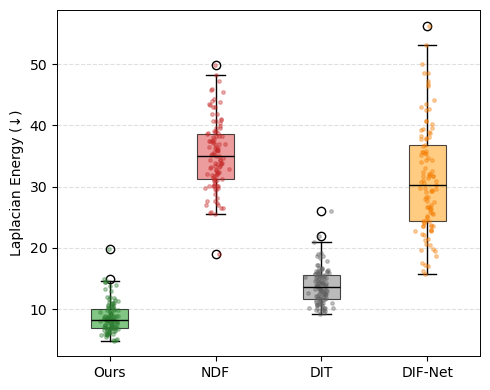}
  \vspace{-6pt}
  \caption{Distribution of discrete Laplacian energy $\|L u\|_2^2$ of vertex displacements across subjects. Lower values correspond to smoother deformation fields.}
  \vspace{-4pt}
  \label{fig:Lenrg}
\end{figure}

\begin{figure*}[t]
  \centering
  \includegraphics[width=1.0\linewidth]{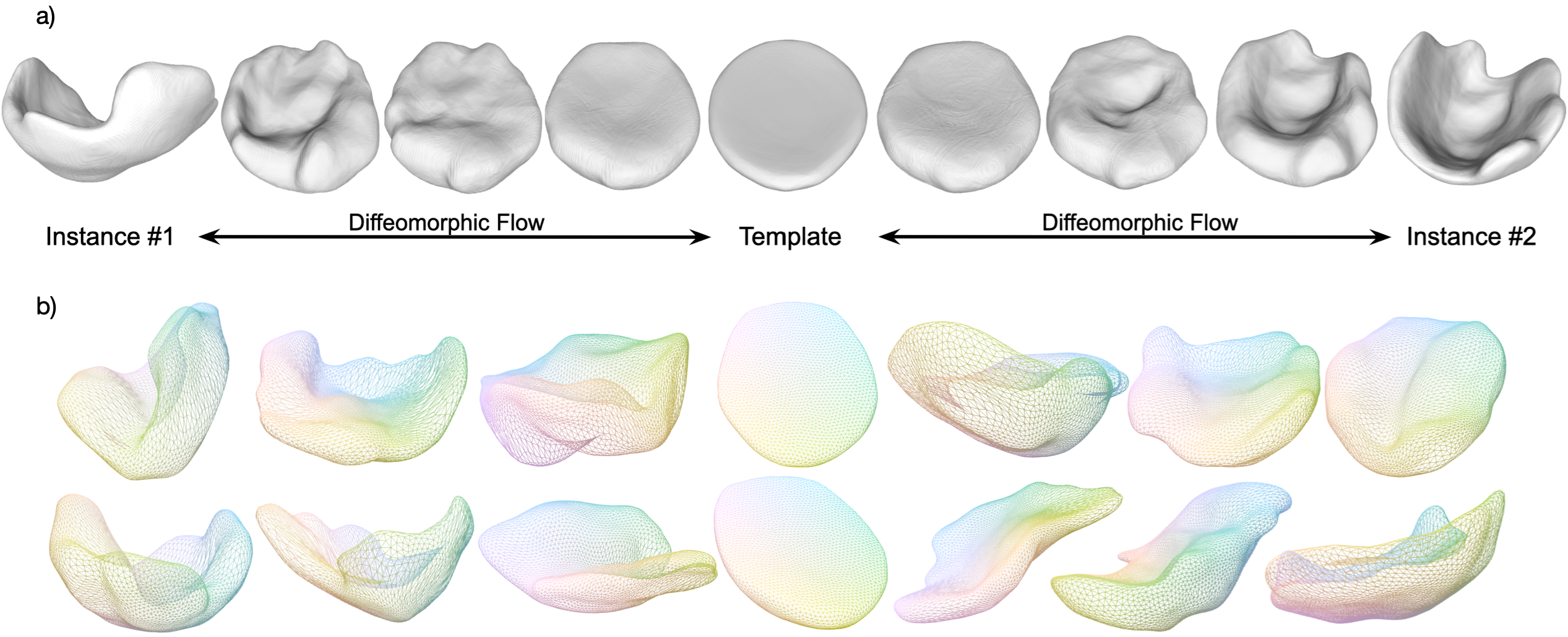}
  \caption{Dense pointwise correspondences via the learned implicit template. (a) Two subject instances mapped to the canonical template through the diffeomorphic flow. (b) ACVD‐remeshed surfaces (5 k vertices) with vertex colors propagated from the template. Consistent colors across subjects visually demonstrate stable, smooth correspondences.}
  \label{fig:flow}
\end{figure*}

At the fine threshold ($\tau=0.01$), our model attains the highest precision, recall, and $\mathrm{F}_1$ scores, reflecting more accurate localization of the zero-level set and reduced bias in the reconstructed surface.
Performance differences diminish for $\tau \geq 0.03$, where all methods approach saturation near $\mathrm{F}_1=100\%$, consistently with the shared SDF sampling and reconstruction regime.

\subsection{Diffeomorphism and Distortion Analysis}
Table~\ref{tab:diff} summarizes diffeomorphism related metrics computed on tetrahedral embeddings. 
Our model attains the lowest flip rate (4.0\%) and smallest $\lVert\!\log|\det\nabla\Phi|\!\rVert_1$, 
which quantifies the deviation of local volume changes from unity, indicating more consistent volume preservation compared to baselines.
The reduced cycle-consistency error $\|\Phi^{-1}(\Phi(x)) - x\|_2$ further confirms improved invertibility of the learned deformation flow.
These gains, though moderate in magnitude, are systematic across all measures and highlight the stabilizing effect of volumetric regularization and Jacobian control during integration. 
See supplementary material for full metric distributions.

Table~\ref{tab:distortion} evaluates geometric distortion using volumetric, areal, and edge stretch ratios, as well as symmetric Dirichlet~\cite{schreiner2004inter, smith2015bijective} and conformal distortion energies~\cite{levy2023least}. All these measures are dimensionless and equal to~1 for perfect volume-, area-, or length-preserving mappings, except SymDir, which achieves the isometric minimum of~6.
Our method consistently yields lower mean and maximum distortion, with the symmetric Dirichlet energy reduced from 13.75 to 8.02 and the conformal distortion from 10.96 to 9.74 relative to the strongest baseline model. The improvements correspond to more isotropic and well-conditioned deformations throughout the placenta volume, ensuring uniform local scaling rather than collapse or elongation along preferred directions.

Fig.~\ref{fig:Lenrg} reports the discrete Laplacian energy of vertex displacements $|L u|_2^2$, which measures the spatial curvature of the deformation field. Our model exhibits markedly lower energy and variance, confirming that the biharmonic regularization effectively suppresses high-frequency oscillations while maintaining global smoothness of the deformation field.
Together, these results demonstrate that incorporating volumetric samples and interior regularization improves diffeomorphic stability and reduces internal distortion beyond what surface-based constraints alone achieve.

\begin{table}[t]
\centering
\footnotesize 
\setlength{\tabcolsep}{1.2pt}
\renewcommand{\arraystretch}{1.1}
\caption{Maximum distortion per instance, averaged across subjects. Smaller values correspond to more isotropic, low‐distortion mappings. SymDir and Conformal report symmetric Dirichlet and condition‐number–based distortions.}
\begin{tabular}{lccccc}
\toprule
 & Vol$(\downarrow)$ & Areal$(\downarrow)$& Edge$(\downarrow)$ & SymDir$(\downarrow)$ &Conformal$(\downarrow)$ \\
\midrule
DIF-Net~\cite{deng2021deformed} & 4.36 & 2.8 &2.43 & 12.91 & 11.47 \\
DIT~\cite{zheng2021deep} & 2.02 & \textbf{0.99} & \textbf{1.13} & 8.48 & 10.89\\
NDF~\cite{Sun_2022_CVPR} & 4.17 & 2.26 & 2.32 & 13.75 & 10.96\\
\midrule
Ours & \textbf{1.94} & 1.17 & 1.19 & \textbf{8.02} & \textbf{9.74}\\
\bottomrule
\end{tabular}
\label{tab:distortion}
\end{table}

\subsection{Dense Correspondences}
Fig.~\ref{fig:flow} visualizes dense pointwise correspondences established through the learned implicit template space.
Each template vertex is assigned a unique color and mapped to individual placenta instances via the inverse flow $\Phi_i^{-1}$.
For consistency across subjects, surfaces are isotropically remeshed to 5\,k vertices using ACVD algorithm~\cite{valette2004approximated}. The resulting color distributions remain highly consistent across subjects, qualitatively indicating smooth and bijective mappings. This observation aligns with the quantitative diffeomorphism results in Table~\ref{tab:diff}, where our model achieves the lowest flip rate and cycle‐consistency error, confirming fold‐free, near‐invertible deformations. Such dense correspondences enable cross‐subject comparison and direct transfer of labels or intensity via a unified template space.

\begin{figure}[t]
  \centering
  \includegraphics[width=0.75\linewidth]{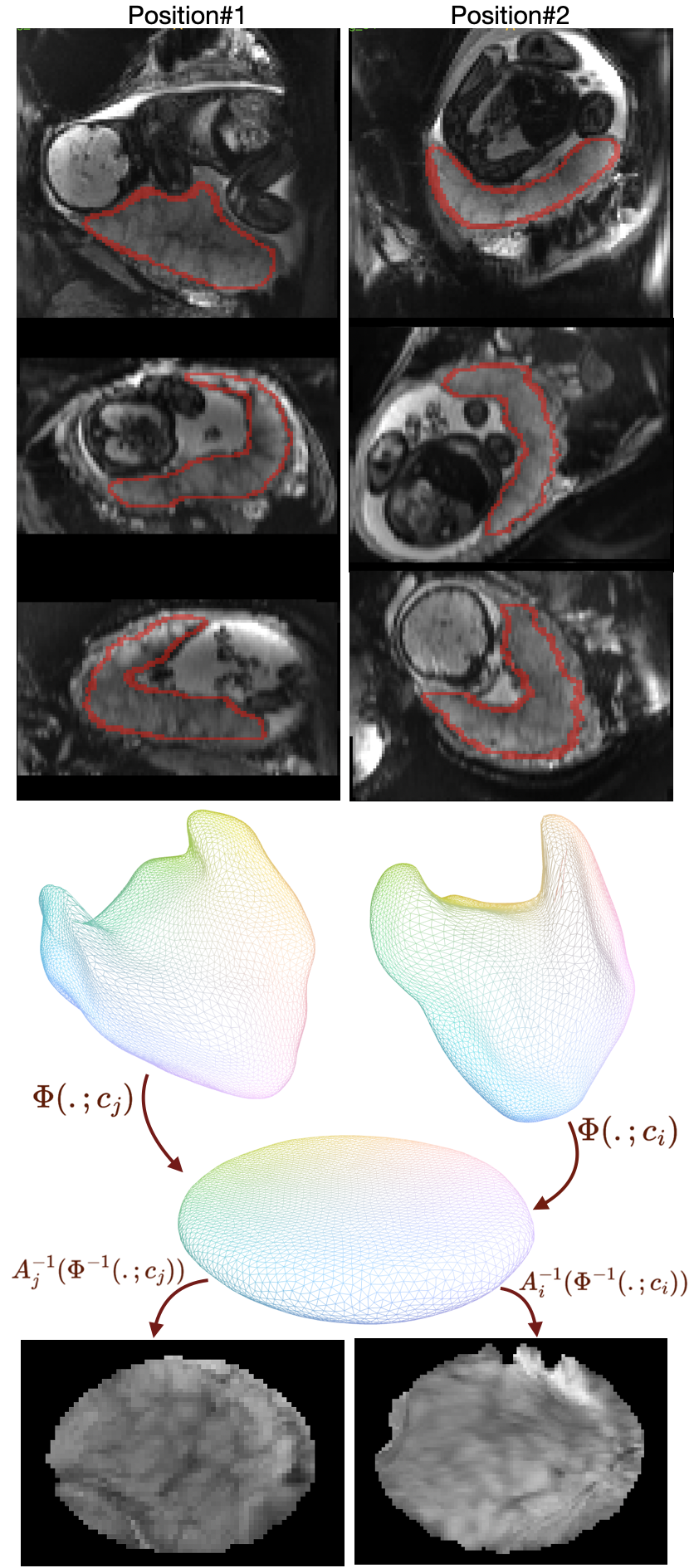}

  \caption{Pose‐invariant flattening and intensity mapping across two maternal positions of the same subject. 
The placenta in each position is mapped diffeomorphically to the canonical template via $\Phi(\cdot;c_i)$ and $\Phi(\cdot;c_j)$, producing consistent, topology‐preserving flattened intensity maps.}
  \label{fig:pose}
\end{figure}
\subsection{Intensity Mapping and Flattening}
The learned diffeomorphic flows enable direct transfer of voxelwise MRI intensities to the canonical template space without additional registration or optimization. Once trained, each placenta volume is mapped to the template domain by the inverse flow $\Phi_i^{-1}$ (Eq.~\ref{eq:intensity_pullback}). 
Fig.~\ref{fig:pose} illustrates this process on two scans of the same subject acquired in different maternal positions. 
Despite large changes in global pose and organ orientation, each placenta is mapped smoothly and bijectively to the common template, and flattened intensity images are generated in a single step without any iterative optimization or inter-volume registration. This contrasts with classical mesh-based methods~\cite{abulnaga2021volumetric, miao2017placenta}, which require solving a volumetric PDE for each instance.

\begin{figure}[b]
  \centering
  \includegraphics[width=1\linewidth]{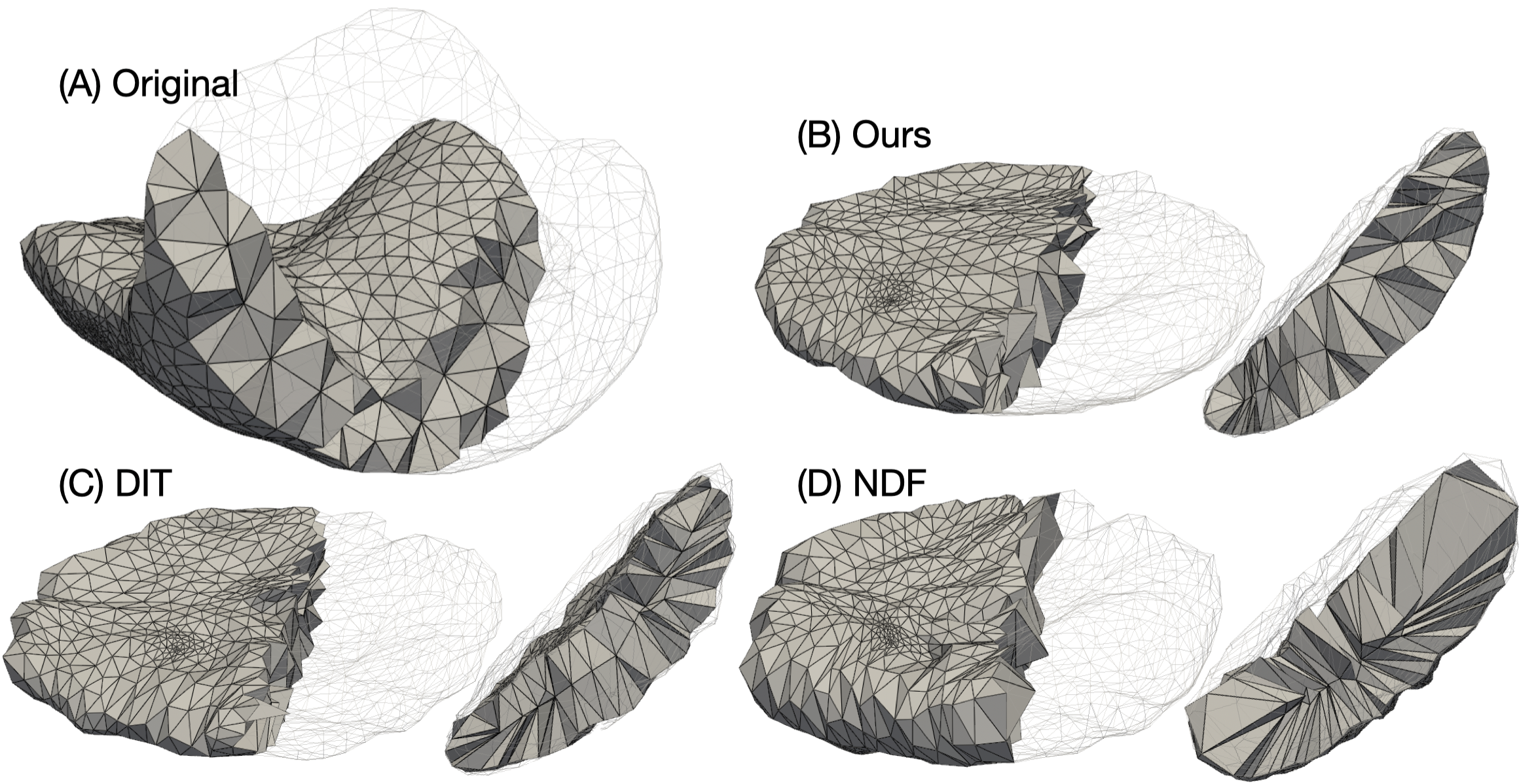}

  \caption{Volumetric flattening. Given the same tetrahedral placenta mesh (upper left panel), our method produces a coherent, uniformly compressed volumetric flattening. 
DIT and NDF maintain surface alignment but distort the interior, exhibiting radial compression and anisotropic stretching of tetrahedra toward a mid-surface layer.}
  \label{fig:tets}
\end{figure}

\begin{figure}[t]
  \centering
  \includegraphics[width=1\linewidth]{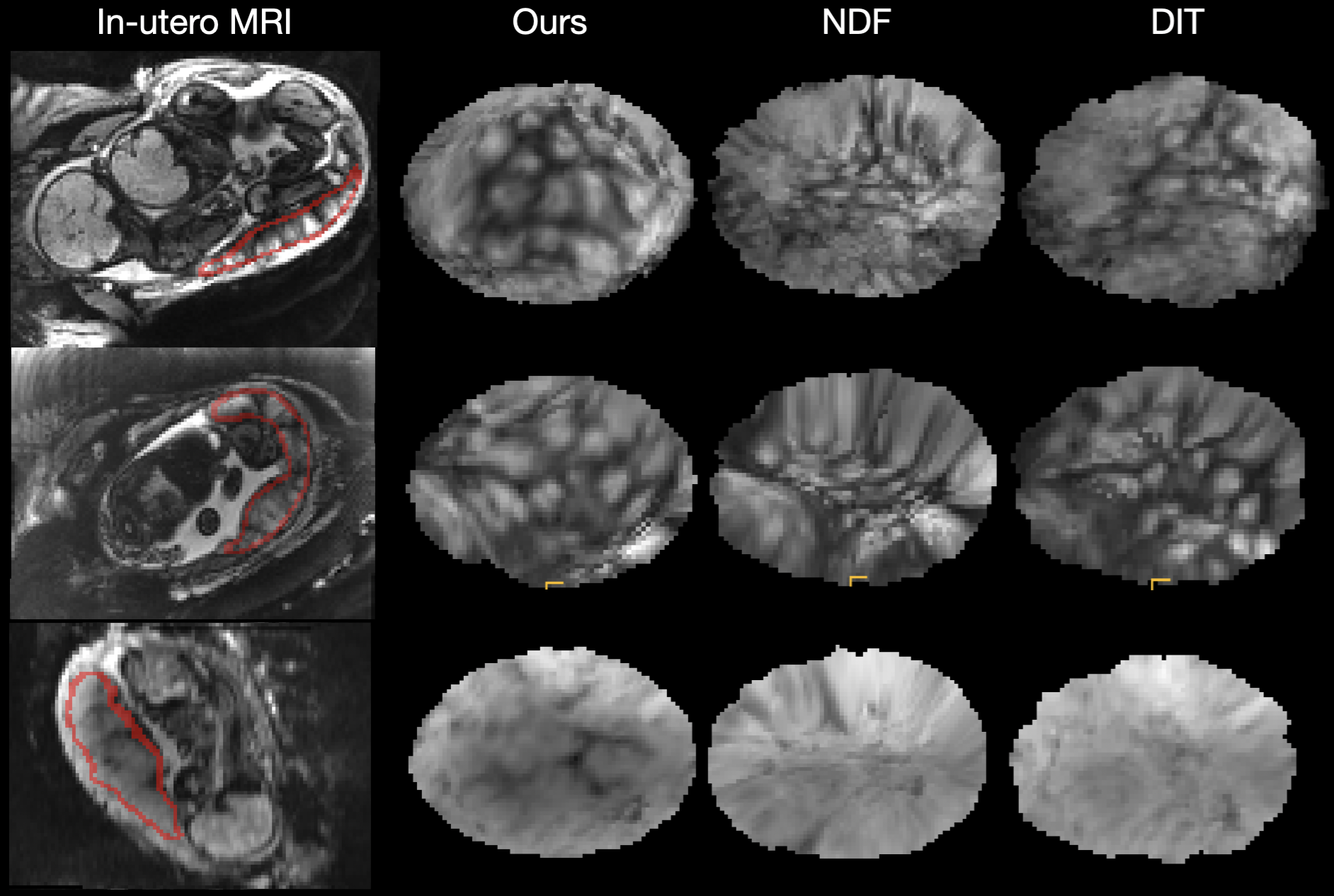}

  \caption{Flattened intensity maps obtained via volumetric pullback. Our method preserves interior geometry, yielding coherent anatomical texture in the flattened domain, cotyledons and vascular structure
  are immediately apparent in our flattened images, whereas NDF and DIT cause intensities to be interpolated across collapsed surface to mid-surface layers. }
  \label{fig:flat}
\end{figure}

Fig.~\ref{fig:tets} shows the deformation of the tetrahedral embedding used for intensity sampling under each model. Our method produces a coherent volumetric compression in which tetrahedra maintain shape quality and follow a consistent diffeomorphic trajectory toward the template. In contrast, DIT and NDF lack interior supervision; their flows are driven almost entirely by surface-level SDF gradients, causing the tetrahedra inside the placenta to collapse toward a mid-thickness layer and form stratified bands between the maternal and fetal surfaces. This radial interior compression leads to anisotropic stretching and loss of volumetric separation between tissue layers. As a result, when intensities are pulled back through these distorted embeddings, voxels near the mid-surface inevitably interpolate across collapsed layers, producing the streaking, blurring, and spoke-like artifacts seen in Fig.~\ref{fig:flat}.   

In Fig.~\ref{fig:flat}, all flattened images are generated at the same voxel resolution as the original in-utero MRI, and the same slice is compared across methods. By preserving full volumetric coherence, our method produces flattened intensity images with clear structural detail. The cotyledons, characterized by their honeycomb pattern of small hyperintense lobules, remain delineated, and the vascular architecture is preserved without radial smearing. Such spatial clarity is essential for downstream evaluation of placental health, oxygenation dynamics, and perfusion heterogeneity, and is obtained here in a single forward pass with no instance-specific optimization.


\subsection{Ablation Study}
We evaluate the contribution of each regularization term in Eq.~\ref{eq:9} using a cumulative ablation. Table~\ref{tab:ablation} reports key geometric and volumetric metrics.

Adding the displacement term $\mathcal{L}_d$ reduces large global warps and leads to improved surface reconstruction compared to $\mathcal{L}_{\mathrm{rec}}$ alone, reflected in lower Chamfer distance and reduced distortion. Introducing the Jacobian constraint $\mathcal{L}_{\mathrm{Jac}}$ increases local injectivity but, in isolation, may over-constrain the flow, resulting in higher flip rate and elevated volumetric distortion; this confirms that a sign constraint alone is insufficient without accompanying smoothness control. Finally, incorporating the bi-harmonic regularizer $\mathcal{L}_{\mathrm{biH}}$ yields the best overall performance, sharply reducing flip rate and Laplacian energy while maintaining surface accuracy. 

\begin{table}[t]
\centering
\footnotesize
\setlength{\tabcolsep}{2.5pt}
\renewcommand{\arraystretch}{1.1}
\caption{Ablation of loss terms on the Placenta dataset.}
\begin{tabular}{lccccc}
\toprule
Model & CD-L2$\downarrow$ & FlipRate$\downarrow$ & logDet–L1$\downarrow$ & SymDir$\downarrow$ & LapE $\downarrow$ \\
\midrule
$\mathcal{L}_{rec}$         & 0.13 & 13.75 & 0.69 & 10.99 & 26.34 \\
w/ $\mathcal{L}_{d}$ & 0.22 & 10.92& \textbf{0.61} & 9.08 & 22.88 \\
w/ $\mathcal{L}_{\text{Jac}}$ & 0.12 & 14.76& 0.80& 12.07 & 33.05 \\
w/ $\mathcal{L}_{\text{biH}}$ (ours)                  & \textbf{0.12} & \textbf{4.04}& 0.69 & \textbf{8.02} & \textbf{8.72} \\
\bottomrule
\end{tabular}
\label{tab:ablation}
\end{table}

\section{Applications and Limitations}
\label{sec:app}
The proposed model provides a unified representation for both anatomical shape analysis and volumetric intensity mapping of the placenta. 
Once trained, the learned diffeomorphic flow enables rapid population‐level alignment, groupwise flattening, and cross‐subject correspondence without additional registration or optimization. 
This facilitates downstream applications such as statistical atlas construction, quantitative comparison of functional MRI signals, and regional analysis of oxygenation or perfusion across gestational age. 
The volumetric regularization further allows faithful interpolation within the organ interior, supporting intensity‐based analyses that were previously limited to mesh surfaces.

While our model achieves high geometric fidelity and volumetric consistency, several limitations remain. Similar to other implicit methods, inference currently requires latent optimization which may limit throughput compared to fully feed‐forward architectures. 
The current formulation assumes a single canonical template; extending to multi‐template or hierarchical atlases could better capture anatomical subtypes and twin‐specific variations. 
Future work will address these extensions and investigate integration with dynamic MRI sequences for spatiotemporal analysis of placental function.

\section{Conclusions}
\label{sec:conclude}
We introduced a volumetrically consistent implicit model that jointly learns a shared placenta template and diffeomorphic flows mapping individual instances into a common coordinate space. 
By enforcing volumetric regularization and pathwise Jacobian constraints, the method achieves smooth, fold-free, and topology-preserving deformations throughout the organ interior. 
The resulting model provides accurate surface reconstruction, dense correspondences, and faithful intensity mapping in a single end-to-end framework, unifying geometric and functional placenta analysis. 
Beyond this application, the formulation offers a general strategy for learning diffeomorphic implicit representations of deformable volumetric anatomy, with potential extensions to other organs and spatiotemporal imaging modalities.
\subsection*{Acknowledgment} This work was supported in part by the MIT Postdoctoral Fellowship Program for Engineering Excellence; the National Institutes of Health (NIH), including NICHD grants R01HD114338 and R21HD106553, and NIH grants R01EB017337, R01AG064027, and UM1MH130981; the MIT Health and Life Sciences Collaborative (HEALS); and the Chou Family Transformative Research Fund. 

{
    \small
    \bibliographystyle{ieeenat_fullname}
    \bibliography{main}
}
\clearpage
\setcounter{page}{1}
\maketitlesupplementary
\renewcommand{\thesection}{\Alph{section}}
\setcounter{section}{0}

\section{Overview}
\label{sec:ovr}
This supplementary document provides additional technical details, experiments, and visualizations supporting our main paper. In Sec.~\ref{sec:Exp}, we describe the MRI dataset and outline the preprocessing steps used to extract surfaces, generate signed-distance samples, and construct tetrahedral embeddings. We then provide implementation specifics, including network architectures, training procedures, and evaluation settings.

\section{Experimental Details}
\label{sec:Exp}

\subsection{Datasets and Preprocessing}

\begin{figure}[b]
  \centering
  \vspace{-4pt}
  \includegraphics[width=0.9\linewidth]{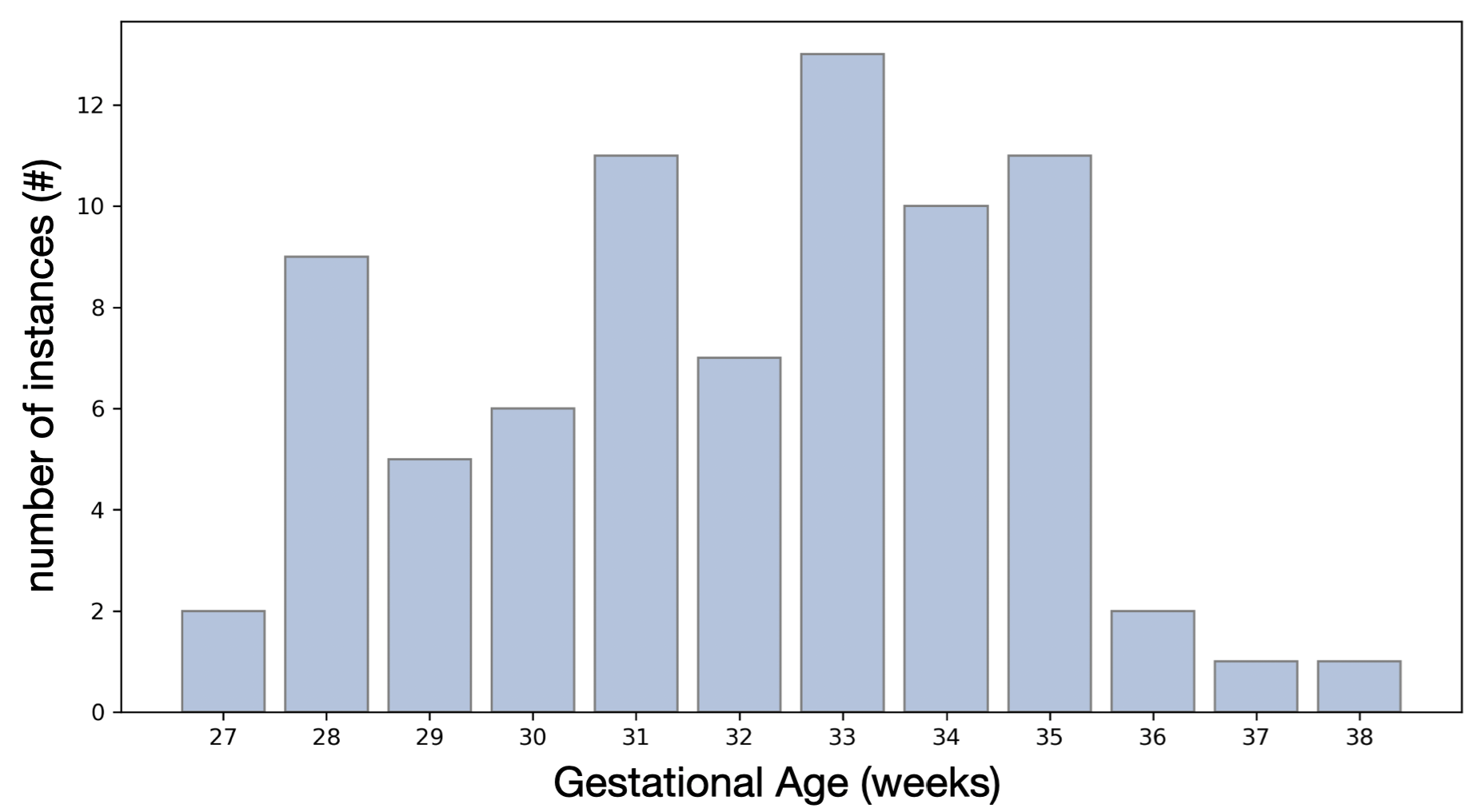}
  \vspace{-6pt}
  \caption{Distribution of gestational ages for our dataset of EPI BOLD placenta scans.}
  \vspace{-4pt}
  \label{fig:age}
\end{figure}

Our primary dataset consists of 111 EPI BOLD MRI scans acquired from 78 pregnant subjects on a 3T Siemens Skyra scanner (single-shot GRE–EPI, 3 mm isotropic voxels, interleaved slice acquisition, TR = 5.8–8 s, TE = 32–36 ms, FA = 90\degree). The cohort includes 60 singleton and 18 monochorionic twin pregnancies. Among singleton cases, 33 subjects were scanned in two maternal positions (supine and left lateral), providing 66 distinct placenta shapes, while the remaining 27 subjects were scanned in a single position. Gestational ages range from 27 to 38 weeks, covering mid-to-late pregnancy (Fig.~\ref{fig:age}). We used 91 placentas for training and 21 placentas for testing, ensuring that no subjects overlap between sets when multiple scans were available for the same pregnancy. All placentas were manually segmented by an expert; small holes were filled, and only the largest connected component was retained before meshing. The resulting volumetric meshes contain 6708$\pm$1813 tetrahedra and 2801$\pm$731 surface triangles. 

\begin{figure}[t]
  \centering
  \includegraphics[width=1.0\linewidth]{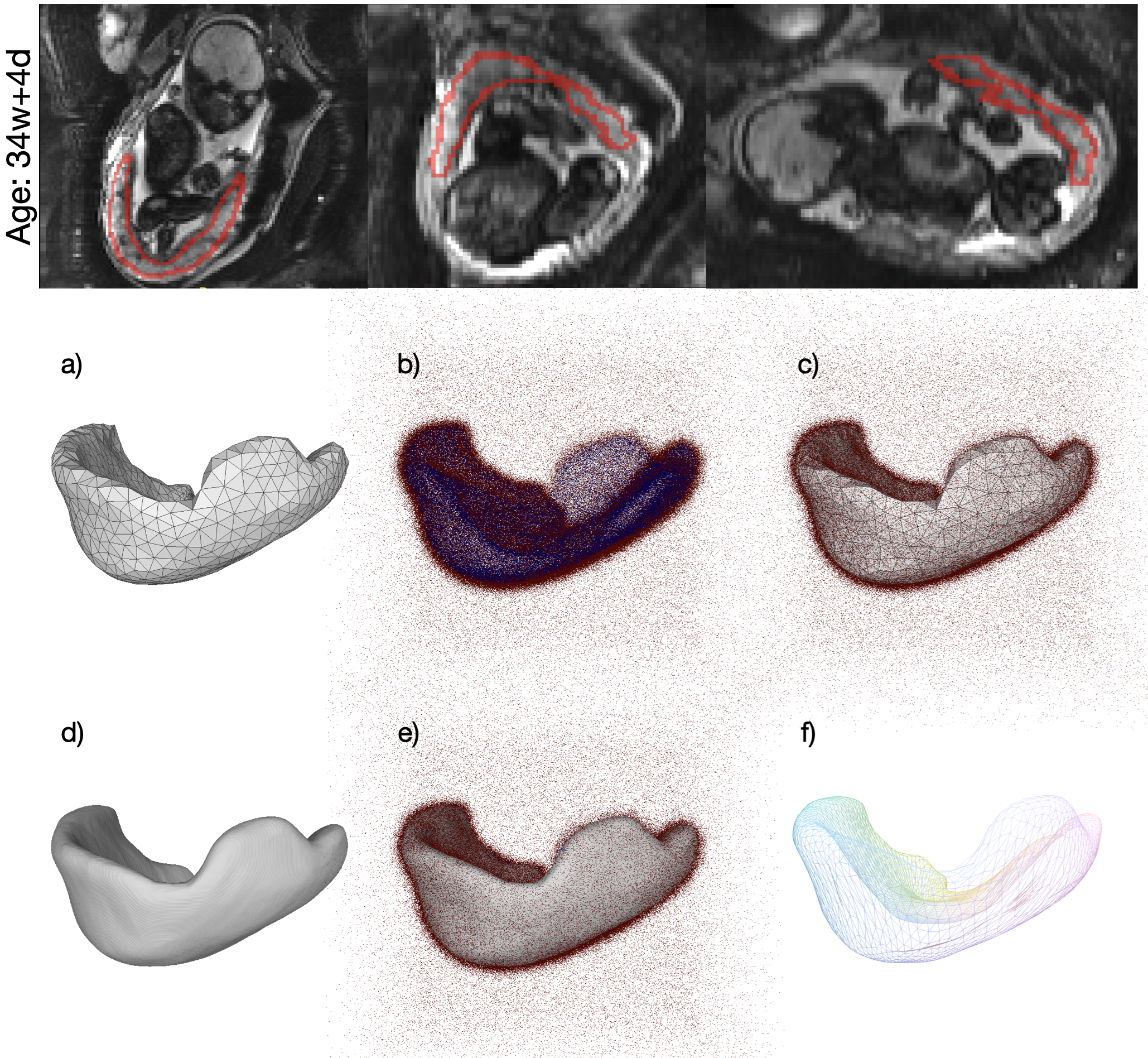}
  \caption{Preprocessing and model outputs for a representative instance with gestational age of 34w+4d (a) Initial surface extracted from the manual segmentation using Marching  Cubes (b) Dense signed-distance samples generated using depth-based nearest-surface queries (red: outside; blue: inside)(c) Overlay of the SDF samples with the initial mesh; a random subset of 5k points is used for training (d) Reconstructed implicit surface produced by our template-conditioned SDF decoder; since the representation is continuous, the surface can be extracted at arbitrarily high resolution independently of the input mesh resolution (e) Overlay of the reconstructed surface with the input SDF samples, illustrating geometric consistency (f) Dense pointwise correspondences obtained from the learned diffeomorphic flow, mapping the subject into the template space.}
  \label{fig:exp}
\end{figure}

\noindent\textbf{Preprocessing.} For each segmentation volume, a surface mesh was first extracted using the Marching Cubes algorithm~\cite{lorensen1998marching}. Meshes were normalized to the unit cube and reoriented such that the first principal axis aligned with the \textit{z}-axis to ensure a consistent anatomical “upright” pose.
Signed-distance samples were generated using depth-based nearest-surface queries. For each case, we sampled points near the surface, producing a mixture of positive (outside), negative (inside), and surface-normal samples (Fig.~\ref{fig:exp}). In addition, a voxelized SDF grid at resolution was computed to provide spatially dense supervision. Tetrahedral embeddings were constructed from the cleaned volumetric meshes and used exclusively for volumetric regularization and geometric evaluation, not for supervision of reconstruction loss.

\subsection{Architecture Details}
\paragraph{Implicit Template Decoder.}
The implicit template $\mathcal{T}_\theta$ is parameterized as a MLP similar to DeepSDF/DIT~\cite{park2019deepsdf, zheng2021deep}. The network takes a 3D point coordinate as input and passes it through five 256-width layers with ReLU activations, weight normalization, and a dropout probability of 0.05. A skip connection is inserted at the 4th layer as in DeepSDF. The final layer uses a hyperbolic tangent activation to produce a bounded SDF value: 3→256→256→256→256→256→1.
This continuous implicit representation allows surfaces to be reconstructed at arbitrary resolution, independent of the resolution of the voxel grid segmentation mesh.

\begin{figure}[b]
  \centering
  \includegraphics[width=1.0\linewidth]{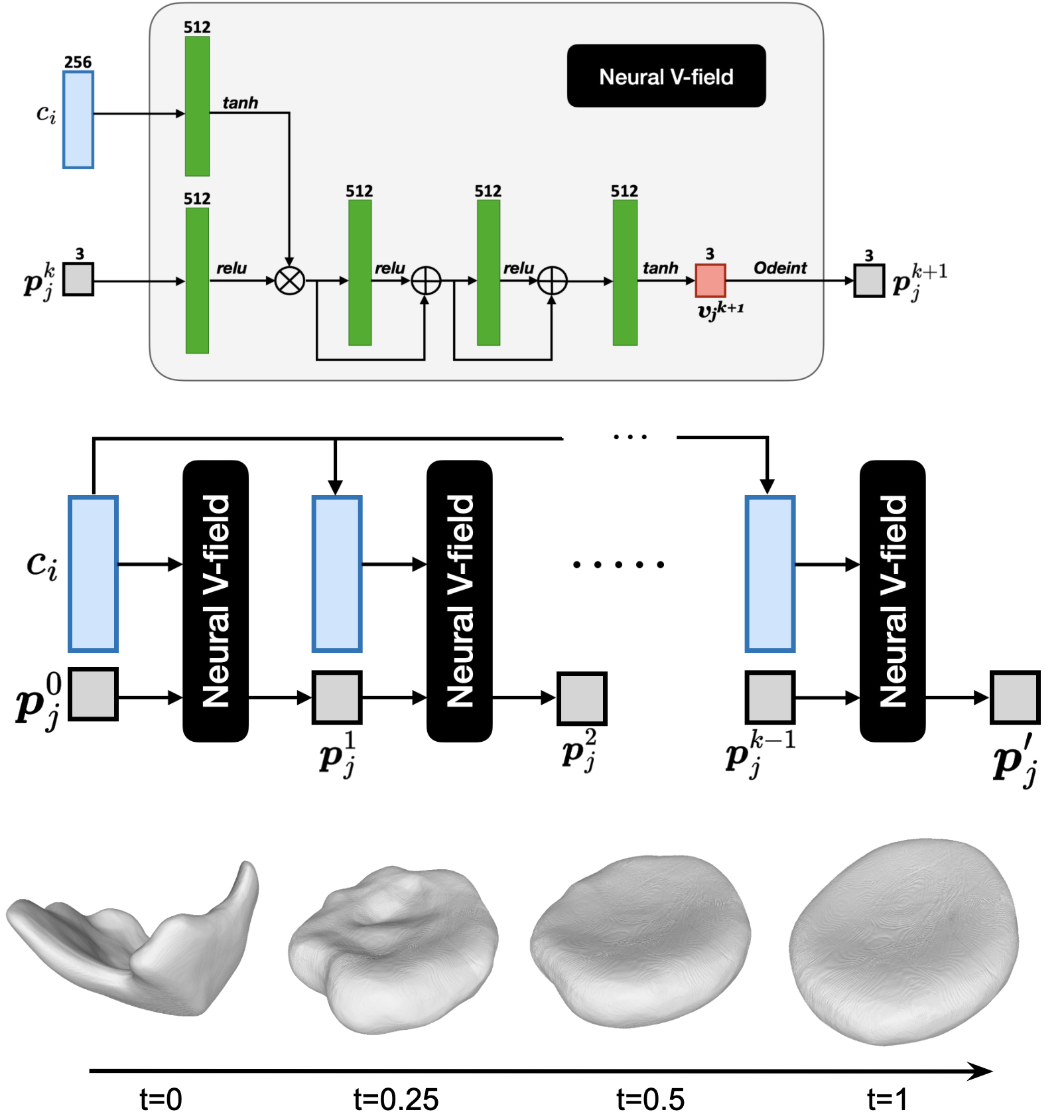}
  \caption{Neural ODE–based deformation module. Top: a velocity field is predicted by a residual MLP from the current point position and latent code $c_i$, and integrated with an ODE solver to produce the updated position. Middle: The full deformation $\Phi_i$ is obtained by concatenating four NODE blocks, where the output of block $k$ serves as the input to block $k+1$. Bottom: Visualization of the progressive deformation steps, illustrating the smooth, monotonic trajectory that warps each placenta instance into the learned template space.}
  \label{fig:net}
\end{figure}

\noindent\textbf{Deformation module.} We adopt a Neural ODE–based warping module following Neural Mesh Flow~\cite{gupta2020neural,Sun_2022_CVPR}. Given a point $p^{0}_j $ sampled from the instance’s SDF, the deformation $\Phi_i$ is defined as a composition of four concatenated NODE blocks, each integrating a learned velocity field over $t \in [0,1]$. Each velocity field is estimated by a fully-connected residual network with hidden width 512: (3+256)→512→512→512→3, with ReLU activations in intermediate layers and tanh at the output to keep displacements within the normalized domain of $[-1, 1]$ . The latent code $c_i$ is concatenated to the input of every NODE block (Fig.~\ref{fig:net}).

\noindent\textbf{Volumetric Embedding and Regularization.} To enforce topological correctness and smooth volumetric deformation, we construct a tetrahedral embedding $\mathcal{M}_i = (V_i, K_i)$ for each instance from the cleaned segmentation mesh. During training, we sample 2{,}000 tetrahedra per instance and penalize negative Jacobian determinants of $\Phi_i$ using a hinge loss with $\epsilon_{\mathrm{raito}}=0.03$ and weight $\lambda_{\mathrm{Jac}}=0.5$. To promote smooth interior deformations, a biharmonic penalty is applied to 1{,}500 interior vertices using the symmetric normalization scheme and weight $\lambda_{\mathrm{biH}}=0.2$. These volumetric terms complement the SDF loss but do not supervise surface geometry.
\subsection{Training and Inference Details}
We jointly optimize network parameters and latent codes using Adam for 2000 epochs. The learning rate for the deformation and template networks is initialized at $5\times10^{-4}$ and reduced by a factor of 0.5 every 500 epochs, while latent codes use a separate optimizer with learning rate $1\times10^{-3}$. Each batch contains four instances, and for every instance we randomly sample 5{,}000 SDF points. SDF values are clamped at a truncation band of 0.1. Latent codes are regularized by $\lambda\left\|\boldsymbol{c}_i\right\|_2^2$ with $\lambda=10^{-4}$ and constrained to lie within a unit $\ell_2$-ball. During inference, the latent codes are optimized for 2400 iterations using an Adam optimizer with a learning rate of $5\times10^{-2}$.

\section{More Experiments}
\label{sec:Vis}

\begin{figure*}[t]
  \centering
  \includegraphics[width=1.0\linewidth]{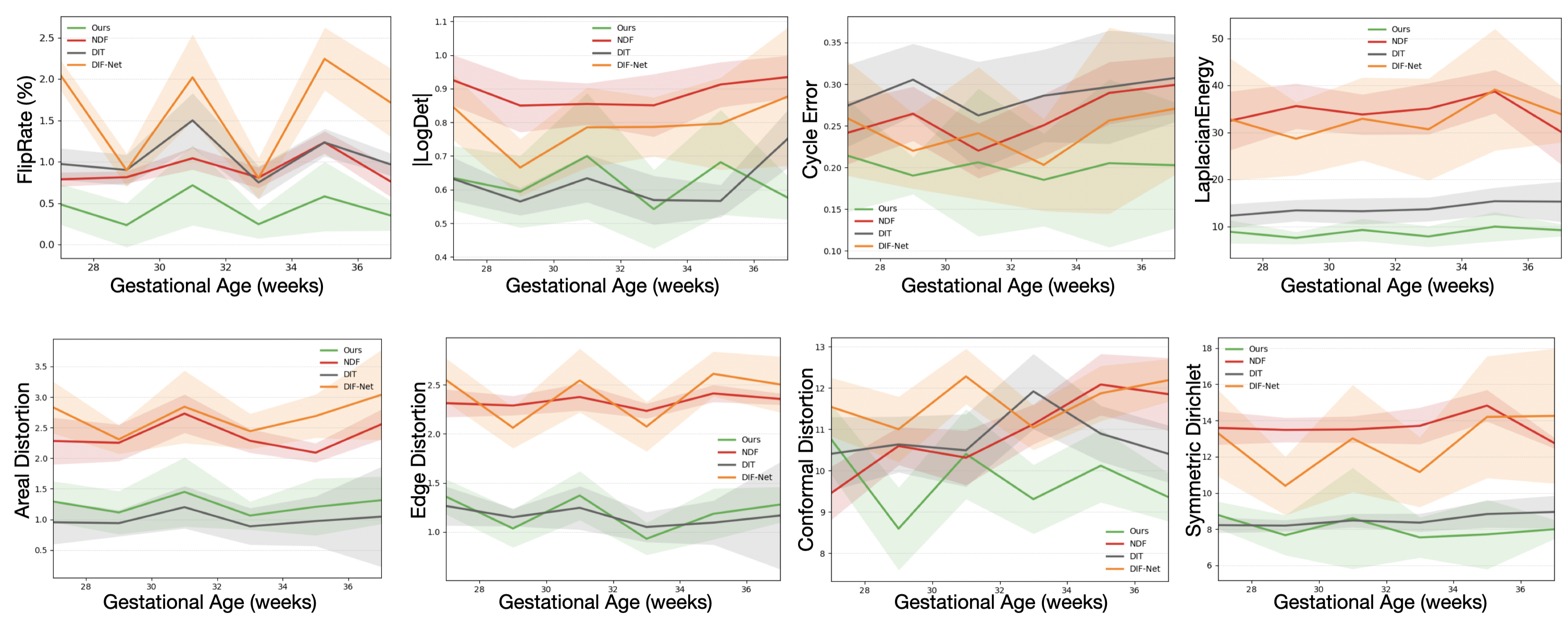}
  \caption{Distribution of diffeomorphism and distortion metrics across gestational age.}
  \label{fig:diff}
\end{figure*}

\noindent\textbf{Diffeomorphism Analysis.} We report the full distribution of diffeomorphic metrics including flip rate, absolute log-determinant, cycle-consistency error, and Laplacian energy sorted by increasing gestational age (GA). For each GA bin, values are averaged over all placentas within that interval, with shaded regions indicating variability across instances. Across all gestational ages, our method consistently achieves lower flip rates, smaller $|\mathrm{LogDet}|$, reduced cycle error, and lower Laplacian energy compared to baseline methods, demonstrating more stable and smoother volumetric deformations throughout late pregnancy (Fig.~\ref{fig:diff}).

\noindent\textbf{Distortion Analysis.} We evaluate geometric distortion using volumetric, areal, and edge stretch ratios, together with symmetric Dirichlet energy~\cite{schreiner2004inter,smith2015bijective} and conformal distortion~\cite{levy2023least}. All distortion measures are dimensionless and attain a value of 1 for perfect volume-, area-, or length-preserving mappings (the symmetric Dirichlet energy attains its isometric minimum at 6). Across all metrics, our method exhibits lower distortion than baseline methods (Fig.~\ref{fig:diff}). These improvements result in deformations that are better-conditioned and more regularized, reducing directional distortion throughout the volume and yielding smoother, more volume-preserving behavior.

\end{document}